\documentclass[conference]{IEEEtran}
\IEEEoverridecommandlockouts
\usepackage{cite}
\usepackage{amsmath,amssymb,amsfonts}
\usepackage{algorithmic}
\usepackage{graphicx}
\usepackage{textcomp}
\usepackage{xcolor}

\usepackage[utf8]{inputenc} \usepackage[T1]{fontenc}    \usepackage{hyperref}       \usepackage{url}            \usepackage{booktabs}       \usepackage{amsfonts}       \usepackage{nicefrac}       \usepackage{microtype}      \usepackage{subcaption}
\usepackage{algorithm}
\usepackage{amsmath}
\usepackage{mathabx}
\usepackage{wrapfig}
\usepackage{diagbox}
\usepackage[mathscr]{eucal}
\newsavebox{\bigimage}
\usepackage[shortlabels]{enumitem}
\usepackage{pifont}

\def\BibTeX{{\rm B\kern-.05em{\sc i\kern-.025em b}\kern-.08em
    T\kern-.1667em\lower.7ex\hbox{E}\kern-.125emX}}

\begin{document}

\newcommand{\textbff}[1]{\textcolor{gray}{\textbf{#1}}}

\title{Training-free Heterogeneous Model Merging}


\author{Zhengqi Xu$^1$, Han Zheng$^1$, Jie Song$^1$, Li Sun$^{1*}$, Mingli Song$^1$\\
$^1$Zhejiang University\\
{\tt\small \{xuzhengqi,h.zheng,sjie,lsun,brooksong\}@zju.edu.cn}
}

\maketitle

\renewcommand{\thefootnote}{\fnsymbol{footnote}}
\footnotetext[1]{Corresponding author.}%
\renewcommand{\thefootnote}{\arabic{footnote}}

\begin{abstract}

Model merging has attracted significant attention as a powerful paradigm for model reuse, facilitating the integration of task-specific models into a singular, versatile framework endowed with multifarious capabilities. Previous studies, predominantly utilizing methods such as Weight Average (WA), have shown that model merging can effectively leverage pretrained models without the need for laborious retraining. However, the inherent heterogeneity among models poses a substantial constraint on its applicability, particularly when confronted with discrepancies in model architectures. To overcome this challenge, we propose an innovative model merging framework designed for heterogeneous models, encompassing both depth and width heterogeneity. To address depth heterogeneity, we introduce a layer alignment strategy that harmonizes model layers by segmenting deeper models, treating consecutive layers with similar representations as a cohesive segment, thus enabling the seamless merging of models with differing layer depths. For width heterogeneity, we propose a novel elastic neuron zipping algorithm that projects the weights from models of varying widths onto a common dimensional space, eliminating the need for identical widths. Extensive experiments validate the efficacy of these proposed methods, demonstrating that the merging of structurally heterogeneous models can achieve performance levels comparable to those of homogeneous merging, across both vision and NLP tasks. 
Our code is publicly available at \href{https://github.com/zju-vipa/training_free_heterogeneous_model_merging}{https://github.com/zju-vipa/training\_free\_heterogeneous\_model\_merging}.
\end{abstract}

\section{Introduction}



Deep neural networks have achieved extraordinary success across a spectrum of demanding computer vision and natural language processing tasks, culminating in the development and public release of numerous models, alongside their architectures and pretrained parameters (\textit{e.g.}, Pytorch Hub\footnote{\url{https://pytorch.org/hub/}}, Hugging Hub\footnote{\url{https://huggingface.co/HUB}}). These easily accessible models are meticulously fine-tuned for various tasks, offering considerable convenience to practitioner. However, their utility remains constrained to the specific tasks for which they were initially trained. This limitation presents significant challenges in terms of model storage~\cite{fifty2021efficiently, zhang2021survey} and computational efficiency, particularly as the size of model parameters grows at an unprecedented rate.

Given the plethora of well-trained models across diverse tasks, a prominent research direction in recent years has been to combine multiple task-specific models into a single model endowed with broad capabilities, without the burdensome need for an exhaustive retraining phase. The existing body of literature can be broadly categorized into two schools: \textit{direct weight average}\cite{yu2023dare_language, ainsworth2023git, modelsoups} and \textit{align-then-average}\cite{ainsworth2023git,tatro2020optimizing_neuron_alignment,stoica2024zipit}.  The former method directly averages the network weights of multiple networks to achieve expanded abilities~\cite{ilharco2023editing,yu2023dare_language} or to enhance generalization performance~\cite{modelsoups}. However, such approaches are confined to networks sharing a common segment of the training trajectory (\textit{e.g.}, the same pretrained model), as the pronounced differences in parameter spaces between models trained via entirely disparate trajectories can result in significant performance degradation~\cite{frankle2018lottery}. To relax this assumption, the latter approaches first align the parameter spaces of the models~\cite{ainsworth2023git,tatro2020optimizing_neuron_alignment,Verma2024MergingTT} and then merge the models through weight averaging, relying on the well-established conjecture that most SGD solutions belong to a set whose elements can be permuted such that no performance barrier exists in the linear interpolation between any two permuted elements~\cite{entezari2021role}.

Despite prior research having made notable strides in pretrained model merging devoid of any training, these all presuppose that the pretrained models exist within the homogeneous architecture, thereby constraining their utility in the face of structurally heterogeneous models. To the best of our knowledge, only a few works~\cite{luo2019knowledge} endeavor to fuse structurally heterogeneous models, but they necessitate a costly retraining phase. The challenge of training-free model merging for heterogeneous models remains largely unexplored, owing to the formidable difficulties posed by model heterogeneity. Specifically, models may differ not only in layer depth but also in layer width, rendering their parameter spaces incompatible for alignment through the element-wise one-to-one mapping employed in existing methods.



In this work, we present a pioneering model merging framework designed to tackle the aforementioned challenge, concentrating on two dimensions of architectural heterogeneity: \textit{depth heterogeneity} and \textit{width heterogeneity}. Specifically, with respect to depth heterogeneity where the number of layers differs, we observe that adjacent layers of the model often exhibit similar representations \cite{cka}, and the input and output of consecutive intermediate layers can be substituted with fewer layers, or even a single layer \cite{efficient_layer_compression}. Accordingly, we introduce a depth-heterogeneous model merging algorithm, which initially partitions the deeper model into multiple segments, with each segment comprising layers exhibiting similar representations. We ensure that the number of segments corresponds to the number of layers in the shallower model, thereby addressing the inconsistency in the number of layers. Regarding width heterogeneity, prior approaches necessitate that the two models share identical widths (i.e., dimensions) in order to establish a one-to-one mapping between the neurons. In contrast, we introduce an elastic neuron zipping algorithm that constructs mapping matrices to project weights of differing widths onto a common width, thus circumventing the need for identical widths. Extensive experiments are conducted to investigate the efficacy of these proposed methods, demonstrating that the proposed heterogeneous model merging can achieve performance comparable to those of homogeneous merging, on both vision and NLP tasks.


In summary, the principal contributions of this paper are:
(1) We explore, for the first time, the challenges inherent in model merging, specifically addressing how to merge models in structurally heterogeneous settings, encompassing both width and depth heterogeneity.
(2) We propose a novel model merging framework that facilitates effective merging in scenarios characterized by both width and depth heterogeneity.
(3) Extensive experimental validation showcases the efficacy of our framework across a range of tasks and model architectures. 

\section{Related Work}

\label{sec:related_work}


\textbf{Direct weight average.} Weight averaging \cite{modelsoups} is a widely used model merging technique that constructs the merged model by averaging parameters. Task Arithmetic \cite{ilharco2023editing} employs a predefined scaling factor to differentiate the significance of various models. Fisher Merging \cite{matena2022fisher_merging} performs weighted parameter fusion, where the weights are determined using the Fisher information matrix \cite{fisher}. RegMean \cite{jin2023dataless} adeptly addresses model merging by optimizing a linear regression problem with closed-form solutions. TIES-Merging \cite{yadav2023ties-merging} resolves task conflicts in \cite{ilharco2023editing} by pruning low-magnitude parameters, rectifying sign disagreements, and merging parameters with consistent signs in isolation. DARE \cite{yu2023dare_language} further mitigates parameter interference from previous approaches by randomly dropping delta parameters and rescaling the remaining ones.  

\textbf{Align-then-average.} Git Re-Basin \cite{ainsworth2023git} and Neuron Alignment \cite{tatro2020optimizing_neuron_alignment} permute models by evaluating the similarity between their weights or activations. REPAIR \cite{jordan2023repair} enhances the precision of Git Re-Basin by calculating the correlation between intermediate layer feature activations and incorporating multiple batch normalization layers into the network. OTFusion \cite{otfusion_singh2020model} introduces a permutation-based approach grounded in optimal transport theory, utilizing the Wasserstein distance, where neuron associations facilitate the one-shot fusion of pre-existing models with identical depths. Several studies \cite{imfeld2024transformerfusion, Verma2024MergingTT} extend these methods to accommodate Transformer-based architectures, though substantial performance degradation persists without fine-tuning. Zipit! \cite{stoica2024zipit} addresses intra-model merging, aligning all models within the same basin by "zipping" redundant features both within and across models. Furthermore, MuDSC \cite{xu2024training} proposes the simultaneous alignment of models in both weight and activation spaces.

\section{Methodology}

\subsection{Preliminaries}

We first review the methodology for merging models in homogeneous architectures. Consider a model $\mathcal{L}$ as a collection of layers $L_{i}\in\mathcal{L}$, each has a set of parameters (\textit{e.g.}, $\boldsymbol{W}_i$, $b_i$ for a linear layer). The task of merging two models $\mathcal{L}^A$ and $\mathcal{L}^B$ involves fusing their parameters into a new model $\mathcal{L}^*$, such that $\mathcal{L}^*$ preserves the accuracy of $\mathcal{L}^A$ and $\mathcal{L}^B$ on their respective original tasks. When $\mathcal{L}^A$ and $\mathcal{L}^B$ are fine-tuned from the same checkpoint, several studies \cite{Izmailov2018AveragingWL,modelsoups} have demonstrated that merging them is as straightforward as averaging their weights. For example, if $L_i$ represents a linear layer and $\boldsymbol{W}_i^A,\boldsymbol{W}i^B \in \mathbb{R}^{n_in{i-1}}$, where $n_i$ denotes the dimension of the $i$-th layer, the new weight matrix $\boldsymbol{W}_i^*$ is simply expressed as
\begin{equation} 
\label{eq:avg} 
\boldsymbol{W}_i^* = \frac{1}{2} \boldsymbol{W}_i^A + \frac{1}{2} \boldsymbol{W}_i^B 
\end{equation}
However, when $\mathcal{L}^A$ and $\mathcal{L}^B$ are not fine-tuned from the same checkpoint, Eqn. \ref{eq:avg} generally yields random accuracy. To address this issue, a body of work \cite{ainsworth2023git,jordan2023repair,stoica2024zipit} has found that permuting the feature space of one model to align with that of the other before averaging significantly recovers lost accuracy. Specifically, following the general framework of prior studies \cite{stoica2024zipit}, let $\boldsymbol{P}_i^A$ and $\boldsymbol{P}_i^B$ represent the permutation matrices that align the output of layer $L_i^A$ and $L_i^B$ to the same space, with $\boldsymbol{P}_i^A, \boldsymbol{P}_i^B \in \mathbb{R}^{n_in_i}$. For each layer, we can apply
\begin{equation}
\boldsymbol{W}_i^*=\boldsymbol{P}^A_i\boldsymbol{W}_i^A(\boldsymbol{P}^A_{i-1})^{-1}+\boldsymbol{P}^B_i\boldsymbol{W}_i^B(\boldsymbol{P}^B_{i-1})^{-1}
\end{equation}
Here, we permute not only the output space of $\boldsymbol{W}i^A$ and $\boldsymbol{W}i^B$, but also their input spaces to reverse the permutation from the previous layer (hence the use of pseudo-inverse matrices $(\boldsymbol{P}{i-1}^A)^{-1}$ and $(\boldsymbol{P}{i-1}^B)^{-1}$).

Let $\boldsymbol{f}_i^A$ and $\boldsymbol{f}_i^B$ represent the feature vectors of the $i$-th layer for each model, where $\boldsymbol{f}_i^A, \boldsymbol{f}_i^B \in \mathbb{R}^{n_im_i}$, and $m_i$ denotes the feature dimension. The search for optimal $\boldsymbol{P}_i^A$ and $\boldsymbol{P}_i^B$ can be formulated as the following objective:
\begin{equation}
\label{objective}
    \mathop{\arg \max}_{\boldsymbol{P}_i^A,\boldsymbol{P}_i^B} \sum_{i=1}^{|\mathcal{L}*|} Sim_{f}(\boldsymbol{P}_i^A\boldsymbol{f}_i^A,\boldsymbol{P}_i^B\boldsymbol{f}_i^B).
\end{equation}
Here, $Sim_{f}(\cdot, \cdot)$ computes the sum of similarities between features at corresponding indices in the two sets of feature vectors. Cosine similarity is commonly employed as $Sim_{f}(\cdot, \cdot)$.

\subsection{Depth-heterogeneous Merging}

Eqn. \ref{objective} reveals that, for depth-homogeneous models, our optimization goal is to maximize the aggregate feature similarities across each layer. However, for depth-heterogeneous models, this formulation is not applicable, as the layers of the two models cannot be directly aligned in a one-to-one manner. Fortunately, prior research has demonstrated that adjacent layers often exhibit similar representations, and the functionality of multiple layers can be effectively replaced by a single independent layer. Drawing inspiration from this, we align the layers of the two models by segmenting the deeper model and treating consecutive layers with analogous representations as a unified segment. Specifically, we assume Model A is deeper than Model B, and partition the layers of Model A into a set of segments $\mathcal{S}^A$. Let ${f}_{ij}^A$ represent the feature from the $j$-th layer of the $i$-th segment $S_i^A$ of Model A, and $\boldsymbol{f}_i^B$ denote the features of the $i$-th layer of Model B. The objective for depth-heterogeneous merging is thus formulated as:
\begin{equation}
\label{eq:depth_objective}
    \mathop{\arg \max}_{\boldsymbol{P}_{ij}^A,\boldsymbol{P}_{ij}^B,S_i^A} \sum_{i=1}^{|\mathcal{L}^B|} \sum_{j=1}^{|S_i^A|} Sim_{f}(\boldsymbol{P}_{ij}^A\boldsymbol{f}_{ij}^A,\boldsymbol{P}_{ij}^B\boldsymbol{f}_i^B),
\end{equation}
where $\boldsymbol{P}_{ij}^A$ and $\boldsymbol{P}_{ij}^B$ are the reprojection matrices corresponding to the $j$-th layer of the $i$-th segment for the Model A and Model B, respectively.

To simplify the problem, we reduce it to a two-step optimization process. First, we determine the segments $\mathcal{S}^A$ of Model A. Then, within each segment $S_i^A$, we sequentially optimize for $\boldsymbol{P}_{ij}^A$ and $\boldsymbol{P}_{ij}^B$, based on $\boldsymbol{f}_{ij}^A$ and $\boldsymbol{f}_{i}^B$, respectively. This gives rise to two key questions: \textit{1) How should we merge $\mathcal{S}^A$ and $\mathcal{L}^B$?} and \textit{2) How do we determine the segments $\mathcal{S}^A$?}

For the first question, to simplify the notation, we assume that we are merging the segment $S^A = \{L_1^A, L_2^A, \dots, L_l^A\}$ of Model A with the layer $L^B$ of Model B. Here, $\boldsymbol{W}_j^A$ and $\boldsymbol{W}^B$ denote the weights of the respective models, while $\boldsymbol{P}_j^A$ and $\boldsymbol{P}_j^B$ refer to the permutation matrices in Eqn. \ref{eq:depth_objective}. Let $\boldsymbol{x}$ represent the input data. Considering that the feature map $\boldsymbol{f}_l$ behave similarly to a linear map (up to a scaling factor $\alpha$) on the line interpolation between $\boldsymbol{W}_l^A$ and $\boldsymbol{W}_l^B$, \textit{i.e.} $\alpha \boldsymbol{f}_l^A+(1-\alpha)\boldsymbol{f}_l^B~\propto~\alpha \boldsymbol{W}_l^A+(1-\alpha)\boldsymbol{W}_l^B$ \cite{NEURIPS2023_llfc}.  We aim to derive a reasonable form of weight averaging from feature averaging for depths-heterogeneous merging. The fused features $\boldsymbol{f}^*$ can be viewed as a synthesis of the features from $S^A$ and those from $L^B$, \textit{i.e.}
\begin{equation}
\begin{aligned}
\label{eq:fused_fea}
    \boldsymbol{f}_l^* &= \boldsymbol{P}_l^A\boldsymbol{W}_l^A(\boldsymbol{P}_{l-1}^A)^{-1}\boldsymbol{f}_{l-1}^* + \boldsymbol{P}_1^B\boldsymbol{W}^B\boldsymbol{x} \\
     &=\boldsymbol{P}_l^A\boldsymbol{W}_l^A(\boldsymbol{P}_{l-1}^A)^{-1}\boldsymbol{P}_{l-1}^A\boldsymbol{W}_{l-1}^A(\boldsymbol{P}_{l-2}^A)^{-1}\dots \boldsymbol{P}_1^A\boldsymbol{W}^A\boldsymbol{x} \\
     &+\boldsymbol{P}_l^B\boldsymbol{I}(\boldsymbol{P}_{l-1}^B)^{-1}\boldsymbol{P}_{l-1}^B\boldsymbol{I}(\boldsymbol{P}_{l-2}^B)^{-1}\dots \boldsymbol{P}_1^B\boldsymbol{W}^B\boldsymbol{x},
\end{aligned}
\end{equation}
where $\boldsymbol{I}$ is the identity matrix. The aforementioned factor $\alpha$ is incorporated into the permutation matrix $\boldsymbol{P}_l$. According to the second term of Eqn. \ref{eq:fused_fea}, $L^B$ can be extended as $S^B = \{L_1^B, L_2^B, \dots, L_l^B\}$, where $L_1^B=L^B$ and $L_i^B(i>1)$ can be regarded as a layer with a weight of $\boldsymbol{I}$. Therefore, merging $S^A$ and $L^B$ can be formulated as merging the layers in $S^A$ and $S^B$ one by one through weight averaging and thus the weights of merged model are derived as
\begin{equation}
\label{eq:hetero_merge}
\begin{aligned}
    \boldsymbol{W}_1^*&=\boldsymbol{P}^A_1\boldsymbol{W}_1^A(\boldsymbol{P}^A_{0})^{-1}+\boldsymbol{P}^B_1\boldsymbol{W}^B(\boldsymbol{P}^B_{0})^{-1}\\
    \boldsymbol{W}_2^*&=\boldsymbol{P}^A_2\boldsymbol{W}_2^A(\boldsymbol{P}^A_{1})^{-1}+\boldsymbol{P}^B_2\boldsymbol{I}(\boldsymbol{P}^B_{1})^{-1} \\
    &\cdots, \\
    \boldsymbol{W}_{l}^*&=\boldsymbol{P}_{l}^A\boldsymbol{W}_{l}^A(\boldsymbol{P}_{l-1}^A)^{-1}+\boldsymbol{P}_{l}^B\boldsymbol{I}(\boldsymbol{P}_{l-1}^B)^{-1}.
\end{aligned}
\end{equation}
We elaborate in the supplementary materials on the approach to merging residual models with heterogeneous architectures.

\begin{figure}[t]
  \centering
  \includegraphics[width=0.5\textwidth]{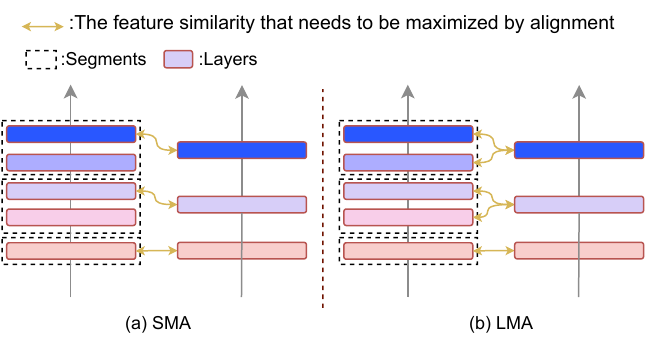}
   \caption{An illustrative diagram of the proposed segment-wise model alignment (SMA) and layer-wise model alignment (LMA) algorithms.}
   \label{fig:align_align}
\end{figure}

For the second question, the objectives can be simplified to finding a set of indices $\mathcal{G}$ such that $S_i^A=\{L_{\mathcal{G}_{i-1}+1}^A,L_{\mathcal{G}_{i-1}+2}^A,...,L_{\mathcal{G}_{i}}^A\}$. In the followings, we present two heuristic algorithms for model alignment (Fig. \ref{fig:align_align}): 1) segment-wise model alignment, and 2) layer-wise model alignment.


\textbf{Segment-wise Model Alignment (SMA).} The goal is to ensure that, after segmentation, the output representation of each segment $S_i^A$ is as similar as possible to the representation of the corresponding layer $L_i^B$ in Model B. To this goal, we firstly compute the pairwise similarity between all layers of Model A and Model B, and then we design a matching algorithm to maximize the similarities between $\boldsymbol{f}_{\mathcal{G}_i}^A$ and $\boldsymbol{f}_{i}^B$:
\begin{equation}
\label{eq:sim_seg}
    \mathop{\arg \max}_{\mathcal{G}} \sum_{i=1}^{|\mathcal{L}^B|} Sim_l(\boldsymbol{f}_{\mathcal{G}_i}^A, \boldsymbol{f}_i^B).
\end{equation}
It is worth noted that, for similartiy function $Sim_l(\cdot, \cdot)$, since the $\boldsymbol{f}_{\mathcal{G}_i}^A$ and $\boldsymbol{f}_i^B$ have not yet been reprojected by the projection matrices, the sum of the similarities of features at corresponding indices cannot be considered as the overall similarity between the feature groups. Therefore, we apply \textit{Centered Kernel Alignment (CKA)} \cite{cka} as 
a proxy to computes the representation similarity between layers because the similarity index is equivalent to $CKA$, \textit{i.e.}~there is  $CKA(\boldsymbol{f}^A,\boldsymbol{f}^B)=CKA(\boldsymbol{P}^A\boldsymbol{f}^A,\boldsymbol{P}^B\boldsymbol{f}^B)$ and thus we are able to measure the similarity between layers prior to alignment.

\textbf{Layer-wise Model Alignment (LMA).} The previous methods primarily focused on aligning the output features of the $S_i^A$ with the features of $L_i^B$. However, as shown in Eqn. \ref{eq:depth_objective}, alignment also occurs between the internal features of $S_i^A$ and the features of $L_i^B$. Therefore, we propose an alignment method that maximizes global feature similarity, whose objective can be formulated as following:

\begin{equation}
\label{eq:diff_seg}
    \mathop{\arg \max}_{\mathcal{G}} \sum_{i=1}^{|\mathcal{L}^B|}  \sum_{j=i}^{\mathcal{G}_i-1} Sim_l(\boldsymbol{f}_{j}^A, \boldsymbol{f}_i^B).
\end{equation}


The pseudocode of the segment- and layer-wise model alignment algorithms are provided in the supplementary materials.

\subsection{Width-heterogeneous Merging}

\label{sec:width_hetero}

\begin{figure}[t]
  \centering
  \includegraphics[width=0.5\textwidth]{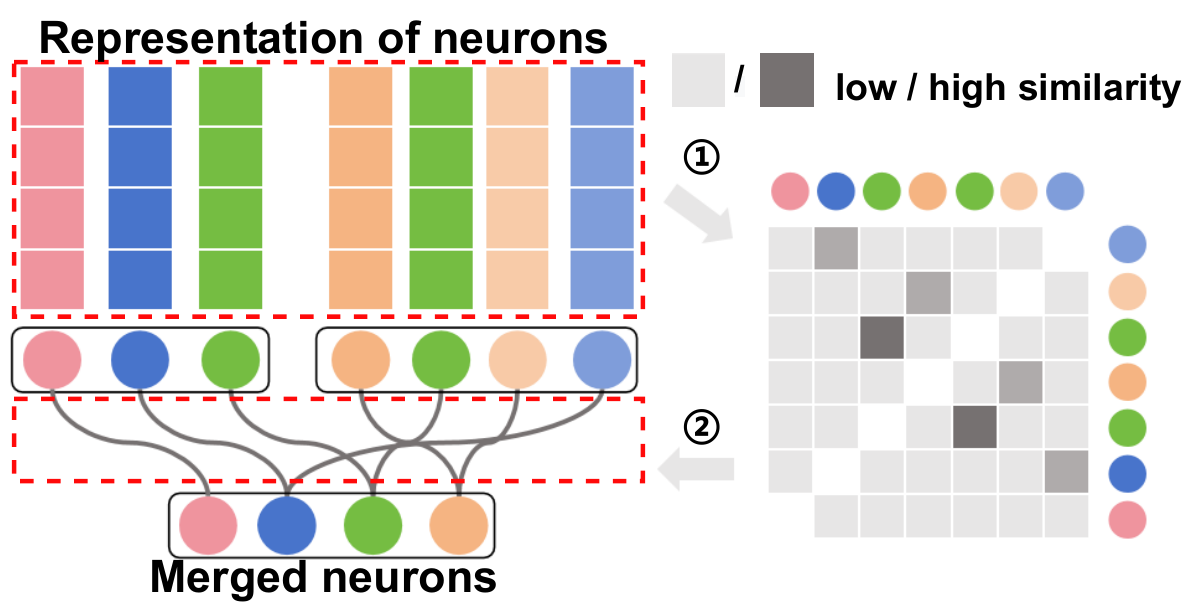}
   \caption{Merging two layers with different width. \textcircled{1} We calculate the similarity for each pair of neurons with their feature vector. \textcircled{2} We merge the most similar neurons according to the similarity until a specified number $r$ of neurons remain, where $r=4$ in the diagram.}
   \label{fig:width_hetero_merge}
\end{figure}

The disparity in width between models is a more prevalent scenario; however, existing methodologies exclusively address the merging of models with identical widths. On one hand, neuron alignment-based techniques \cite{ainsworth2023git, tatro2020optimizing_neuron_alignment} necessitate the establishment of a one-to-one correspondence between neurons of equal quantity. On the other hand, a neuron zip-based approach \cite{stoica2024zipit} has been demonstrated effective solely for models of identical width. In this work, we introduce an elastic neuron zipping algorithm that can accommodate models of arbitrary widths and merge the correlated neurons. Fig. \ref{fig:width_hetero_merge} illustrates the process of merging layers with heterogeneous widths. As shown in Fig. \ref{fig:width_hetero_merge}, similar neurons are merged one by one regardless of the model they belong to, and the number of remaining neurons is constrained by the predefined hyperparameter $r$. In practice, $r$ can be set to the maximum width of the models being merged.

\section{Experiments}

\subsection{Experimental Settings}

\textbf{Datasets}. 
The experiments are performed on both vision and natural language tasks, encompassing the small-scale CIFAR-10/100 \cite{cifar_krizhevsky2009learning}, the large-scale ImageNet~\cite{imagenet}, and the renowned General Language Understanding Evaluation (GLUE) benchmark for natural language comprehension \cite{wang-etal-2018-glue}.

\textbf{Models.} We adopt various commonly used model architectures to demonstrate the to illustrate the versatility of the proposed method. For vision tasks, we merge ResNets \cite{resnet_he2016deep} and VGGs \cite{simonyan2014very} with varying depths and widths. For natural language understanding classification tasks, we investigate Transformer encoder-based masked language models. Specifically, we consider 5 different BERT models, seeds 1 through 5, from the MultiBERTs reproduction \cite{devlin-etal-2019-bert} and each model has 12 layers. To obtain models with different depths, we repeated the even-numbered layers of each model, extending the depth of the models to 17 layers. For each classification task in GLUE, we fine-tune each of the MultiBERTs models with a randomly initialized classification head,including pooling layer and classification layer weights. We keep the head initializations the same across models. 

\textbf{Evaluation}. For the experiments on CIFAR-10/100 and ImageNet, We randomly partition a classification dataset into two non-overlapping sub-classification tasks, trained respective models for each, and subsequently merged the models into one. Then we evaluate performance of merged model with joint accuracy and per-task accuracy. Joint accuracy is the overall accuracy of a model when it is evaluated on all classes within a combined dataset. 
For per-task accuracy, we provided the accuracy of the merged multi-task model on two individual tasks, along with their average performance. 
Each model is trained with a CLIP-style loss \cite{clip_radford2021learning} using CLIP text encodings of the class names as targets. For fair comparisons, we train 3 pairs of models and report the average accuracy. For the experiments on GLUE, we investigate loss barriers between fine-tuned BERT models across 8 different GLUE tasks. We use the loss-barrier defined by Frankle \textit{et.al.} \cite{frankle2018lottery}. 


\subsection{Results of Merging Vision Models}

\begin{table*}[t]
\caption{The experiments for merging depth-heterogeneous/homogeneous models.}
\begin{center}
\resizebox{1.0\linewidth}{!}{
\begin{tabular}{c|cccccccc|cccc}
\toprule
Models          & \multicolumn{8}{c|}{Resnet26+Resnet50}                                                   & \multicolumn{4}{c}{VGG13+VGG19}  \\
\hline
Datasets        & \multicolumn{4}{c|}{CIFAR100(50+50)}                 & \multicolumn{4}{c|}{CIFAR10(5+5)} & \multicolumn{4}{c}{CIFAR10(5+5)} \\
\hline
Methods         & Joint & Avg   & Task A & \multicolumn{1}{c|}{Task B} & Joint  & Avg    & Task A & Task B & Joint  & Avg   & Task A & Task B \\
\midrule
Task A         & 41.67 & 53.00 & 83.05  & \multicolumn{1}{c|}{22.94}  & 48.32  & 66.24  & 96.00  & 36.48  & 47.53  & 63.67 & 94.65  & 32.70  \\
Task B         & 41.18 & 52.61 & 23.03  & \multicolumn{1}{c|}{82.18}  & 48.30  & 64.82  & 33.46  & 96.18  & 47.95  & 62.66 & 30.25  & 95.07  \\
Homo$_{Avg}$       & 18.12 & 24.08 & 23.89  & \multicolumn{1}{c|}{24.26}  & 42.22  & 62.34  & 66.43  & 58.25  & 10.38  & 21.09 & 21.22  & 20.96  \\
Homo$_{Align}$     & 46.92 & 62.95 & 62.94  & \multicolumn{1}{c|}{62.95}  & 63.96  & 87.18  & 87.15  & 87.20  & 48.48  & 79.41 & 79.45  & 79.37  \\
Homo$_{Zip}$       & 55.47 & 67.18 & 67.18  & \multicolumn{1}{c|}{67.17}  & 83.34  & 93.77  & 93.78  & 93.77  & 73.20  & 89.95 & 89.95  & 89.95  \\
\hline
Hetero$_{A.SMA.}$  & 45.96 & 62.98 & \textbf{69.67}  & \multicolumn{1}{c|}{56.28}  & 66.65  & 88.41  & \textbf{90.32}  & 86.50  & 46.83  & 72.24 & 65.44  & \textbf{79.03}  \\
Hetero$_{A.LMA.}$ & \textbf{46.55} & \textbf{63.20} & 68.48  & \multicolumn{1}{c|}{\textbf{57.93}}  & \textbf{67.58}  & \textbf{88.55}  & 88.15  & \textbf{88.95}  & \textbf{48.86}  & \textbf{76.17} & \textbf{73.42}  & 78.93  \\
\hline
Hetero$_{Z.SMA.}$  & 55.84 & 67.80 & \textbf{70.32}  & \multicolumn{1}{c|}{65.28}  & 83.15  & 93.68  & 93.93  & 93.42  & 70.17  & 85.95 & 85.01  & 86.89  \\
Hetero$_{Z.LMA.}$ & \textbf{55.99} & \textbf{68.71} & 69.76  & \multicolumn{1}{c|}{\textbf{67.65}}  & \textbf{83.18}  & \textbf{94.16}  & \textbf{94.77}  & \textbf{93.54}  & \textbf{70.40}  & \textbf{86.15} & \textbf{85.32}  & \textbf{86.98} \\
\bottomrule
\end{tabular}
}
\label{tb:depth_merge}
\end{center}
\end{table*}

\begin{table}[htbp]
\caption{The experiments conducted on ImageNet. The two models are ResNet50 and ResNet34, respectively.}
\begin{center}
\resizebox{1.0\linewidth}{!}{
\begin{tabular}{c|cccc}
\toprule
Methods        & Joint Acc. & Avg. Acc. & Task A & Task B \\
\midrule
Task A        & 36.70      & 39.70     & 72.56  & 6.83   \\
Task B        & 37.28      & 40.32     & 7.07   & 73.56  \\
Homo$_{Avg}$      & 65.25      & 73.06     & 72.56  & 73.56  \\
Homo$_{Align}$    & 27.50      & 32.91     & 32.75  & 33.07  \\
Homo$_{Zip}$      & 32.25      & 36.83     & 36.94  & 36.72  \\
\hline
Hetero$_{A.SMA.}$  & 30.48      &  35.42   & 30.40  & 40.44  \\
Hetero$_{A.LMA.}$ & \textbf{30.49}      & \textbf{35.68}     & \textbf{30.81}  & \textbf{40.56}  \\
\hline
Hetero$_{Z.SMA.}$ & 32.62      & 41.14     & 37.15  & 45.14  \\
Hetero$_{Z.LMA.}$ & \textbf{32.66}      & \textbf{42.06}     & \textbf{37.91}  & \textbf{46.21} \\
\bottomrule
\end{tabular}
}
\label{tb:imnet_merging}
\end{center}
\end{table}

\textbf{Depth-heterogeneous merging}. The results of depth-heterogeneous merging are reported in Tab. \ref{tb:depth_merge} and Tab \ref{tb:imnet_merging}. The tables include results for three categories of methods: 1) The average performance of the models trained on Task A or Task B. 2) The average performance of the merged models for each pair of homogeneous models. 3) The average performance of the merged models for each pair of depth-heterogeneous models. The \textit{Avg} refers to vanilla averaging of weights. The \textit{Align}(\textit{A.}) refers to merging model via alignment-based method \cite{ainsworth2023git}. The \textit{Zip}(\textit{Z.}) refers to merging model via zip-based method \cite{stoica2024zipit}. The \textit{SMA.} refers to aligning depth via \textit{segment-wise model alignment}. The \textit{LMA.} refers to aligning depth via \textit{layer-wise model alignment}. As shown in Tab. \ref{tb:depth_merge}, compared to single-task models, the merged models achieve higher joint accuracy and per-task average accuracy, regardless of whether the merging is depth-homogeneous or depth-heterogeneous. For the proposed depth-heterogeneous merging method, the models merged with depth-heterogeneous architectures not only achieve higher average performance compared to the weight average of depth-homogeneous models but also exhibit similar performance to the models merged with depth-homogeneous architectures. It demonstrates the effectiveness of our proposed depth-heterogeneous merging method. 
Furthermore, by comparing different depth alignment methods, we find that depth alignment based on LMA often achieves better fusion performance, particularly in terms of joint accuracy and average per-task accuracy. In the Section \ref{sec:visual}, we further illustrate the reasons behind this phenomenon through visual analysis. Additionally, we conduct experiments on ImageNet (as shown in Tab. \ref{tb:imnet_merging}) and obtained results similar to those in Tab. \ref{tb:depth_merge}, demonstrating the effectiveness of our method on large-scale datasets.

\textbf{Width-heterogeneous merging}. The results for width-heterogeneous merging are reported in Tab. \ref{tb:width_hetero}. 
We train models with different widths for ResNet26 and ResNet50 and merge models with the same depth but different widths. We report the average performance of single-task models with the same architecture as well as the average performance of the width-heterogeneous merged models. As shown in Tab. \ref{tb:width_hetero}, both the joint accuracy and per-task average accuracy of the models merged using the method proposed in Section \ref{sec:width_hetero} significantly outperform the single-task models, demonstrating the feasibility and effectiveness of merging width-heterogeneous models.

\begin{table}[htbp]
\caption{The results for width-heterogeneous merging.}
\begin{center}
\resizebox{1.0\linewidth}{!}{
\begin{tabular}{ccccc}
\toprule
\multicolumn{1}{c|}{Methods}          & Joint Acc. & Avg. Acc.   & Task A & Task B \\
\midrule
\multicolumn{5}{c}{Original Models}                                     \\
\midrule
\multicolumn{1}{c|}{Resnet26$_{\times8}$}     & 41.28 & 52.66 & 52.60  & 52.73  \\
\multicolumn{1}{c|}{Resnet50$_{\times8}$}     & 41.87 & 53.11 & 53.17  & 53.05  \\
\multicolumn{1}{c|}{Resnet26$_{\times4}$}     & 40.88 & 52.36 & 52.11  & 52.60  \\
\multicolumn{1}{c|}{Resnet50$_{\times4}$}     & 41.09 & 52.21 & 51.96  & 52.46  \\
\multicolumn{1}{c|}{Resnet26$_{\times16}$}    & 41.67 & 53.25 & 53.18  & 53.32  \\
\multicolumn{1}{c|}{Resnet50$_{\times16}$}    & 41.74 & 53.61 & 53.41  & 53.80  \\
\midrule
\multicolumn{5}{c}{Width-hetero Merging}                                \\
\midrule
\multicolumn{1}{c|}{Resnet26$_{\times8,\times4}$}  & 59.19 & 70.79 & 72.19  & 69.39  \\
\multicolumn{1}{c|}{Resnet26$_{\times16,\times8}$} & 65.45 & 76.08 & 77.44  & 74.72  \\
\multicolumn{1}{c|}{Resnet50$_{\times8,\times4}$}  & 60.63 & 72.27 & 73.56  & 70.97  \\
\multicolumn{1}{c|}{Resnet50$_{\times16,\times8}$} & 60.07 & 71.85 & 70.99  & 72.71 \\
\bottomrule
\end{tabular}
}
\label{tb:width_hetero}
\end{center}
\end{table}

\subsection{Results of Merging Language Models}

As shown in Tab. \ref{tb:nlp_merging}, we compare the performance of vanilla averaging, homogeneous model merging, and heterogeneous model merging on the GLUE benchmark. Specifically, we implement the vanilla averaging and homogeneous model merging of BERTs based on the method proposed by Verma \textit{et al}\cite{Verma2024MergingTT}. Subsequently, we employed \textit{layer-wise model alignment} to achieve the merging of depth-heterogeneous BERTs. 
To ensure fairness in the experiments, during homogeneous merging, we merge models finetuned with different random seeds across multiple runs. For heterogeneous merging, we increase the depth of one model in each pair only before finetuning. We investigate the loss barriers and errors of the three methods across 8 tasks. Compared to vanilla averaging, lower loss barriers can be observed for homogeneous model merging and heterogeneous model merging.

\begin{table}[htbp]
\caption{The merging results of GLUE. Bold and gray bold text are used to indicate the best and second-best results.}
\begin{center}
\begin{tabular}{c|cc|cc|cc}
\toprule
Methods & \multicolumn{2}{c|}{Vanilla Averaging} & \multicolumn{2}{c|}{Homo Merging} & \multicolumn{2}{c}{Hetero Merging} \\
\hline
Tasks   & Barrier$\downarrow$          & Error          & Barrier$\downarrow$        & Error       & Barrier$\downarrow$        & Error        \\
\midrule
MNLI-mm & \textbf{0.59}                  & 0.06           & \textbff{0.65}               & 0.09        & 0.67                & 0.07         \\
QQP     & 1.38                  & 0.07           &     \textbf{1.11}            & 0.08        &   \textbff{1.16}            & 0.08         \\
QNLI    & \textbf{0.67}                  & 0.04           & 0.76                 & 0.05        &   \textbff{0.71}            & 0.06         \\
SST-2   & 0.52                  & 0.02           & \textbf{0.38}                & 0.08        & \textbff{0.42}                & 0.10          \\
CoLA    & 1.31                  & 0.16           & \textbff{1.11}                & 0.11        & \textbf{1.05}                & 0.14         \\
STS-B   & 5.04                  & 0.39           & \textbff{4.32}                & 0.32        & \textbf{4.24}                & 0.28         \\
MRPC    & 2.81                  & 0.07           & \textbf{1.87}                & 0.12        & \textbff{1.88}                & 0.12         \\
RTE     & 0.54                  & 0.03           & \textbf{0.44}                & 0.03        & \textbff{0.46}                & 0.06     \\
\bottomrule
\end{tabular}
\label{tb:nlp_merging}
\end{center}
\end{table}

\subsection{Visualization Analysis of Depth Alignment}
\label{sec:visual}

\begin{figure}[t]
	\centering
	\begin{subfigure}[b]{1\linewidth}  
		\centering
		\includegraphics[width = \linewidth]{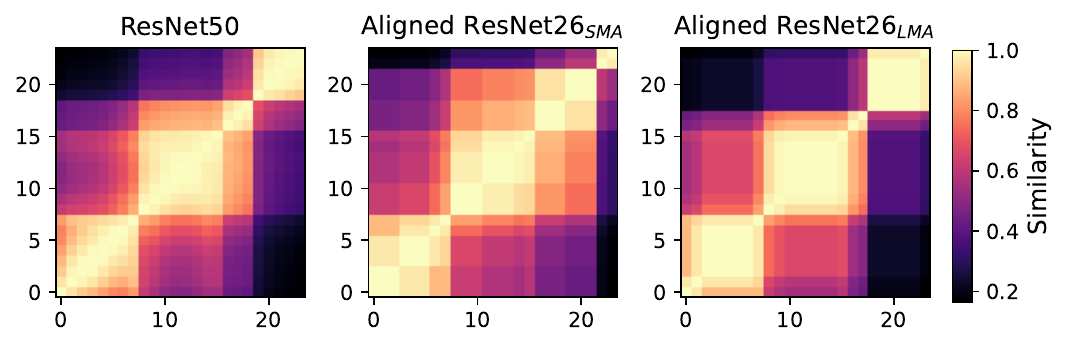}
		\caption{Models v.s. Self}\label{fig:compare2self}
	\end{subfigure}
	\begin{subfigure}[b]{1\linewidth}
		\centering
		\includegraphics[width = \linewidth]{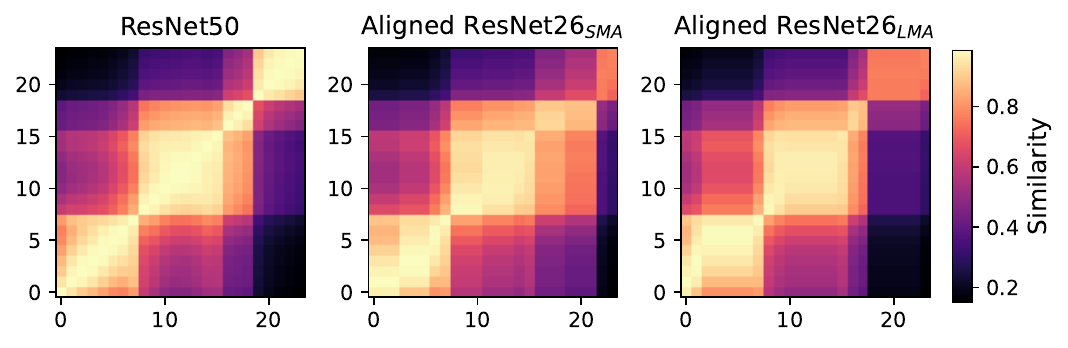}
		\caption{Models \textit{v.s.} ResNet50}\label{fig:compare2origin}
	\end{subfigure}

	\caption{Visualization of representation similarity.}\label{fig:weight_vec}
\end{figure}

Here we demonstrate the differences between the depth alignment methods by visualizing the representational similarity between arbitrary layers of two models. We adopt ResNet50 and ResNet26, both trained on CIFAR100. ResNet26 is extended to match the depth of ResNet50 through different alignment strategies. For clarity, we visualize the representations of each residual block rather than individual layers.

Fig. \ref{fig:compare2self} presents the representational similarity between arbitrary pairs of layers within individual models. 
Although SMA aims to maximize the consistency between the representations of the segments in the aligned deeper model and the layers in the shallower model, it neglects the alignment between intermediate representations within the segments and the final representations of the shallower model's layers. This oversight can lead to potential misalignments. In contrast, LMA ensures that representational shifts occur at similar locations as in the original model, resulting in more appropriate alignment results.

Furthermore, we compare the representational similarity between ResNet50 and aforementioned three models. As shown in Fig. \ref{fig:compare2origin}, the diagonal representational similarity reflects the layer-by-layer similarity between the deep model and the aligned shallow models.  We observe that the optimal alignment for SMA  does not align along the diagonal. In contrast, this issue does not occur with LMA. Moreover, we calculated the mean of the diagonal elements in the similarity matrices corresponding to SMA and LMA in Fig. \ref{fig:compare2origin}, obtaining values of 88.24\% and 89.69\%, respectively. This further demonstrates that the alignment results of LMA are globally superior.

\section{Conclusion}

We present a novel model merging framework designed to address the merging of depth-heterogeneous and width-heterogeneous models. Two heuristic approaches, \textit{segment-wise model alignment} and \textit{layer-wise model alignment}, achieve depth alignment by partitioning the layers of the deeper model into multiple segments, equal in number to the layers of the shallower model. An elastic neuron zipping technique is proposed for the merging of width-heterogeneous models. Through experimental analysis, we demonstrate that the proposed framework for merging heterogeneous models is both feasible and effective across a range of tasks and architectures.

\bibliographystyle{abbrv}
\bibliography{main}


\end{document}


\newcommand{\textbff}[1]{\textcolor{gray}{\textbf{#1}}}

\title{{Training-free Heterogeneous Model Merging}\\ -----\textit{Supplementary Material}-----}


\maketitle

\section{Deep-heterogeneous Merging for ResNet}

In this section, we elaborate on the approach to merging residual models when confronted with heterogeneous architectures. Assume merging the segment $S^A = \{L_1^A, L_2^A, \dots, L_l^A\}$ of Model A with the layer $L^B$ of Model B and all of $L_i^A(i=1,2,\cdots,l)$ and $L^B$ are residual layers which are simply formulated as $\boldsymbol{f}=(\boldsymbol{W}+\boldsymbol{I})\boldsymbol{x}$. Following the proof approach in the case of non-residual layers, we derive the form of weight averaging in residual layers through the averaging of features. In this case, the averaged feature $\boldsymbol{f}^*$ is given by: 

\begin{equation}
\begin{aligned}
\label{eq:res_fused_fea}
    \boldsymbol{f}_l^* &= (\boldsymbol{P}_l^A\boldsymbol{W}_l^A(\boldsymbol{P}_{l-1}^A)^{-1}+\boldsymbol{I})\boldsymbol{f}_{l-1}^* + (\boldsymbol{P}_1^B\boldsymbol{W}^B+\boldsymbol{I})\boldsymbol{x} \\
     &=(\boldsymbol{P}_l^A\boldsymbol{W}_l^A(\boldsymbol{P}_{l-1}^A)^{-1}+\boldsymbol{I})(\boldsymbol{P}_{l-1}^A\boldsymbol{W}_{l-1}^A(\boldsymbol{P}_{l-2}^A)^{-1}+\boldsymbol{I}) \\
     &~~~\dots (\boldsymbol{P}_1^A\boldsymbol{W}^A+\boldsymbol{I})\boldsymbol{x} \\
     &+(\boldsymbol{P}_l^B\boldsymbol{0}(\boldsymbol{P}_{l-1}^B)^{-1}+\boldsymbol{I})(\boldsymbol{P}_{l-1}^B\boldsymbol{0}(\boldsymbol{P}_{l-2}^B)^{-1} + \boldsymbol{I})\\
     &~~~\dots (\boldsymbol{P}_1^B\boldsymbol{W}^B+\boldsymbol{I})\boldsymbol{x},
\end{aligned}
\end{equation}
where $\boldsymbol{0}$ and $\boldsymbol{I}$ are zero matrix and identity matrix respectively. It is noted that we can still extend $L^B$ to $S^B = \{L_1^B, L_2^B, \dots, L_l^B\}$, where $L_1^B = L^B$ but $L_i^B(i>1)$ represents a residual layer with weights set to $\boldsymbol{)}$. This is because residual layers can directly pass features to the next layer via the shortcut connection. Then the weights of merged model are expressed as:

\begin{equation}
\label{eq:res_hetero_merge}
\begin{aligned}
    \boldsymbol{W}_1^*&=\boldsymbol{P}^A_1\boldsymbol{W}_1^A(\boldsymbol{P}^A_{0})^{-1}+\boldsymbol{P}^B_1\boldsymbol{W}^B(\boldsymbol{P}^B_{0})^{-1}\\
    \boldsymbol{W}_2^*&=\boldsymbol{P}^A_2\boldsymbol{W}_2^A(\boldsymbol{P}^A_{1})^{-1}+\boldsymbol{P}^B_2\boldsymbol{0}(\boldsymbol{P}^B_{1})^{-1}\\&=\boldsymbol{P}^A_2\boldsymbol{W}_2^A(\boldsymbol{P}^A_{1})^{-1} \\
    &\cdots, \\
    \boldsymbol{W}_{l}^*&=\boldsymbol{P}_{l}^A\boldsymbol{W}_{l}^A(\boldsymbol{P}_{l-1}^A)^{-1}+\boldsymbol{P}_{l}^B\boldsymbol{0}(\boldsymbol{P}_{l-1}^B)^{-1}\\&=\boldsymbol{P}_{l}^A\boldsymbol{W}_{l}^A(\boldsymbol{P}_{l-1}^A)^{-1}.
\end{aligned}
\end{equation}

\section{Pseudocode}
The pseudocode of the segment- and layer-wise model alignment algorithms are provided in Algorithm \ref{alg:sim_alg} and Algorithm \ref{alg:diff_alg} respectively.

\begin{algorithm}[htbp]
  
  \caption{Segment-wise Model Alignment}
  \label{alg:sim_alg}
\begin{algorithmic}
\REQUIRE  The representation similarity matrix $\boldsymbol{C}$ , where $\boldsymbol{C} \in \mathcal{R}^{|\mathcal{L}^A||\mathcal{L}^B|}$ and $\boldsymbol{C}_{i,j}=Sim_l(\boldsymbol{f}_i^A, \boldsymbol{f}_j^B)$.

\STATE Initialize the matrix $\boldsymbol{T}$ to $\mathbf{0}$.
\FOR{$i=1,2,\dots,|\mathcal{L}^B|$}
    \STATE $\boldsymbol{T}_{i,i} \gets \sum_{j=1}^{i} \boldsymbol{C}_{j,j}$
\ENDFOR
\FOR{$i=1,2,\dots,|\mathcal{L}^B|$}
    \FOR{$j=i+1,i+2,\dots,|\mathcal{L}^A|$}
        \STATE $\boldsymbol{T}_{i,j} \gets max(\boldsymbol{T}_{i,j-1}, \boldsymbol{T}_{i-1,j-1}+\boldsymbol{C}_{j,i})$
    \ENDFOR
\ENDFOR
\STATE $\mathcal{G}_1 \gets 1;\mathcal{G}_{|\mathcal{L}^B|} \gets |\mathcal{L}^A|;i \gets n-1; j \gets m-1$
\WHILE{$i \ge 2$}
    \WHILE{$j \ge i+1~and~\boldsymbol{T}_{i,j}=\boldsymbol{T}_{i,j-1}$}
        \STATE $j \gets j-1$
    \ENDWHILE
    \STATE $\mathcal{G}_i \gets j; i \gets i-1; j \gets j-1$
\ENDWHILE

\STATE \textbf{return} $\mathcal{G}$
\end{algorithmic}

\end{algorithm}

\begin{algorithm}[htbp]
  
  \caption{Layer-wise Model Alignment}
  \label{alg:diff_alg}
\begin{algorithmic}
\REQUIRE  The representation similarity matrix $\boldsymbol{C}$ , where $\boldsymbol{C} \in \mathcal{R}^{|\mathcal{L}^A||\mathcal{L}^B|}$ and $\boldsymbol{C}_{i,j}=Sim_l(\boldsymbol{f}_i^A, \boldsymbol{f}_j^B)$.

\STATE Initialize the matrix $\boldsymbol{T}$ to $\mathbf{0}$.
\FOR{$i=1,2,\dots,|\mathcal{L}^B|$}
    \STATE $\boldsymbol{T}_{i,i} \gets \sum_{j=1}^{i} \boldsymbol{C}_{j,j}$
\ENDFOR
\FOR{$i=1,2,\dots,|\mathcal{L}^B|$}
    \FOR{$j=i+1,i+2,\dots,|\mathcal{L}^A|$}
        \STATE $\boldsymbol{T}_{i,j} \gets max(\boldsymbol{T}_{i,j-1}+\boldsymbol{C}_{j,i-1}, \boldsymbol{T}_{i-1,j-1}+\boldsymbol{C}_{j,i})$
    \ENDFOR
\ENDFOR
\STATE $\mathcal{G}_1 \gets 1;\mathcal{G}_{|\mathcal{L}^B|} \gets |\mathcal{L}^A|;i \gets n-1; j \gets m-1$
\WHILE{$i \ge 2$}
    \WHILE{$j \ge i+1~and~\boldsymbol{T}_{i,j}=\boldsymbol{T}_{i,j-1}+\boldsymbol{C}_{j,i-1}$}
        \STATE $j \gets j-1$
    \ENDWHILE
    \STATE $\mathcal{G}_i \gets j; i \gets i-1; j \gets j-1$
\ENDWHILE

\STATE \textbf{return} $\mathcal{G}$
\end{algorithmic}

\end{algorithm}

\section{More experimental details}
\subsection{Calculating Loss Barrier}
To evaluate the performance on GLUE, we use the loss-barrier defined by  Frankle \textit{et.al.} \cite{frankle2018lottery}
which is defined as the maximum difference between the loss of an interpolation and the average loss of the base models:
\begin{equation}
\label{eq:loss_barrier}
\max_{\lambda}\mathcal{L}(\lambda\theta_A+(1-\lambda)\theta_B)-\frac{1}{2}(\mathcal{L}(\theta_A)+\mathcal{L}(\theta_B)),
\end{equation}
where we compute several interpolations of $\theta_A$ and $\theta_B$, as $\lambda \theta_A+(1-\lambda)\theta_B$, and we use 21 samples evenly spaced between $\lambda=0$ and $\lambda=1$.

\bibliographystyle{IEEEbib}
\bibliography{main}
